\documentclass[submission,copyright,creativecommons]{eptcs}

\providecommand{\event}{TARK 2017} 
\usepackage{breakurl}             
\usepackage{underscore}           

\usepackage[english]{babel} \usepackage[latin1]{inputenc}

\usepackage{amsfonts, amsmath, amssymb, epsfig, graphicx, amsthm}
\usepackage{mathrsfs}
\usepackage{color}
\usepackage{multirow,tabularx}
\usepackage{url}
\usepackage{booktabs}
\usepackage{verbatim}
\usepackage{csquotes}
\usepackage{enumitem}
\usepackage{tikz}
\usepackage{epsfig, graphicx}
\usepackage{changepage}
\usetikzlibrary{arrows}

\MakeOuterQuote{"}

\newtheorem{definition}{Definition}
\newtheorem{thm}{Theorem}
\newtheorem{lemma}[thm]{Lemma}
\newtheorem{proposition}[thm]{Proposition}
\newtheorem{cor}[thm]{Corollary}

\newtheorem{example}{Example}

\newcommand{\G}{\mathcal G}
\newcommand{\act}{\mathcal A}
\newcommand{\B}{\boldsymbol{B}}
\newcommand{\V}{\boldsymbol{V}}

\newcommand{\op}[2]{ \mathsf{op}( #1,#2 ) }
\newcommand{\vis}[2]{ \mathsf{vis}( #1,#2 ) }

\newcommand{\stateset}{ \mathcal{S}}

\newcommand{\N}{\mathcal N}

\newcommand{\ltllogic}{\mathsf{LTL}}

\newcommand{\clpclogic}{\mathsf{CL\text{-}PC}}

\newcommand{\ATL}{\mathsf{ATL}}
\newcommand{\LTL}{\mathsf{LTL}}
\newcommand{\nextop}[1]{ \bigcirc #1 }
\newcommand{\until}[2]{ #1 \mathcal{U} #2}
\newcommand{\atlop}[1]{  \langle\!\langle  #1 \rangle\!\rangle }
\newcommand{\atlstr}[1]{\boldsymbol{\sigma}_{#1}}

\newcommand{\Str}{\boldsymbol{\sigma}}
\newcommand{\outset}[2]{out(#1, #2) }

\begin{document}
\title{Relaxing Exclusive Control in Boolean Games}


\author{Francesco Belardinelli
\institute{IBISC, Universit\'e d'Evry and \\IRIT, CNRS, Toulouse}
\email{belardinelli@ibisc.fr}
\and
Umberto Grandi
\institute{IRIT, University of Toulouse}
\email{umberto.grandi@irit.fr}
\and
Andreas Herzig
\institute{IRIT, CNRS, Toulouse}
\email{andreas.herzig@irit.fr}
\and
Dominique Longin
\institute{IRIT, CNRS, Toulouse}
\email{dominique.longin@irit.fr}
\and
Emiliano Lorini
\institute{IRIT, CNRS, Toulouse}
\email{emiliano.lorini@irit.fr}
\and
Arianna Novaro
\institute{IRIT, University of Toulouse}
\email{arianna.novaro@irit.fr}
\and
Laurent Perrussel
\institute{IRIT, University of Toulouse}
\email{laurent.perrussel@irit.fr}
}

\def\titlerunning{Relaxing Exclusive Control in Boolean Games}
\def\authorrunning{Belardinelli et al.}


\maketitle

\begin{abstract}
In the typical framework for boolean games (BG) each player can change
the truth value of some propositional atoms, while attempting to make
her goal true. In standard BG goals are propositional formulas,
whereas in iterated BG goals are formulas of Linear Temporal
Logic. Both notions of BG are characterised by the fact that agents
have exclusive control over their set of atoms, meaning that no two
agents can control the same atom. In the present contribution we drop
the exclusivity assumption and explore structures where an atom can be
controlled by multiple agents.  We introduce Concurrent Game
Structures with Shared Propositional Control (CGS-SPC) and show that
they account for several classes of repeated games, including iterated
boolean games, influence games, and aggregation games.  Our main
result shows that, as far as verification is concerned, CGS-SPC can be
reduced to concurrent game structures with exclusive control.
This result provides a polynomial reduction for the model checking
problem of specifications in Alternating-time Temporal Logic on CGS-SPC.
\end{abstract}

%

\section{Introduction}\label{sec:intro}

Coalition Logic of Propositional Control $\clpclogic$
was introduced by van der Hoek and Wooldridge \cite{CLPC2015}  as a formal language for reasoning about
capabilities of agents and coalitions in multiagent environments, later extended by the concept of transfer of control \cite{DBLP:journals/jair/HoekWW10}.
In $\clpclogic$, capability is modeled by means of the concept of \emph{propositional control}:
it is assumed that each agent $i$ is associated with a specific finite subset $\Phi_i  $
of the finite set of all atomic variables $\Phi $, which are the variables \emph{controlled} by $i$,
in the sense that $i$ has the ability to assign a (truth) value to each variable in $\Phi_i  $
but cannot change the truth values of the variables in $  \Phi  \setminus \Phi_i$.
Control over variables is assumed to be \emph{exclusive}:
two agents cannot control the same variable, i.e.,
if $i \neq j$ then $\Phi_i \cap \Phi_j = \emptyset$.\footnote{
In $\clpclogic$, it is also assumed that control is \emph{complete},
that is, every variable is controlled by at least one agent (i.e.,
for every $p \in \Phi$ there exists an agent $i$ such that $p \in \Phi_i$).
}
The connection between $\clpclogic$ and Dynamic Logic of Propositional Assignments was explored by Grossi \emph{et al.} \cite{DBLP:journals/jair/GrossiLS15}.

A boolean game BG \cite{Harrenstein2001,Bonzon2006}
is a game in which each player wants to achieve
a certain goal represented by a propositional formula.
Boolean games correspond to the specific subclass of normal form
games in which agents have binary preferences.
They share with $\clpclogic$ the idea that an agent's action consists in affecting
the truth values of the variables she controls.
Just as in there, control over atomic propositions is exclusive in BGs.
More recently, BGs were generalized to iterated boolean games IBGs \cite{GutierrezEtAlIC20015,GHW13}.
In IBGs, the agents' goals are formulas of Linear Temporal Logic $\ltllogic$, and
an agent's strategy determines an assignment of the variables controlled by the agent
in every round of the game.

Gerbrandy was the first to study $\clpclogic$ without exclusive control \cite{Gerbrandy06}.
In his games of propositional control, the value of a variable at the next state is determined by
an outcome function that combines the agents' choices of values for propositional variables.
Gerbrandy's language contains a coalition operator and---just as coalition logic---only allows to reason about what agents and coalitions of agents are able to achieve in a single step.
Importing results from many-dimensional modal logics, Gerbrandy proved that the satisfiability problem is 
decidable when there are at most 2 agents, and undecidable otherwise \cite[Prop.5]{Gerbrandy06}.

The aim of the present paper is to further study models without exclusive propositional control
as a basis for BGs and other game-theoretic approaches.
Specifically, we introduce Concurrent Game Structures with Shared Propositional Control CGS-SPC
and show their relationship with different classes of games studied in literature, including IBGs.
The main result of the paper is that CGS-SPC can be reduced to CGS with
Exclusive Propositional Control CGS-EPC \cite{BH16} by introducing a dummy
agent who controls the value of the shared variables and simulates the transition function.
The reduction is polynomial, showing that the problem of verification of specifications in Alternating-time Temporal Logic on CGS-SPC can be reduced to verification in CGS-EPC.
We also explore the consequences of such results in the problem of finding a winning strategy in games with shared control.

The paper is organized as follows.
Section~\ref{sec:framework} provides the basic definitions of concurrent game structures with exclusive and shared control, as well as introducing the language and the semantic of Alternating-time Temporal Logic.
Section~\ref{sec:games} shows that a number of game structures introduced in the literature can be reconducted to our definition of CGS-SPC.
We then prove our main result in Section~\ref{main_result}, where we reduce the problem of $\ATL^*$ model checking for CGS-SPC to model checking of a translated $\ATL^*$ formula in a CGS-EPC suitably defined.
Section~\ref{applications} discusses the consequences in computational complexity of our main result, and Section~\ref{sec:conclusions} concludes.
%
%
%
%
%
%
%
%

\section{Formal Framework}\label{sec:framework}


In this section we consider two classes of concurrent game structures
with propositional control, suitable for the interpretation of a logic
for individual and collective strategies which is introduced next. The two classes differ in
the type of propositional control: exclusive in the former and
shared in the latter.

\subsection{CGS with Exclusive and Shared Control}

We first present concurrent game structures with exclusive
propositional control CGS-EPC as they have been introduced by Belardinelli and Herzig \cite{BH16}.\footnote{More precisely, the CGS-EPC we consider here as our basic framework correspond to the ``weak'' version defined by Belardinelli and Herzig \cite{BH16}, as opposed to a strong version where $d(i,s) = \mathcal{A}_i$ for every $i \in N$ and
$s \in S$.}
We then generalise them by relaxing the assumption of exclusive control.

%
\begin{definition}[CGS-EPC]\label{def:CGS-PC} A \emph{concurrent game structure with exclusive propositional control} is a
tuple $\G=\langle N, \Phi_1, \dots,$ $\Phi_n, S, d, \tau \rangle$, where:
\begin{itemize}
\item $N = \{1, \dots, n \}$ is a set of \emph{agents};
\item $\Phi=\Phi_1\cup \dots \cup \Phi_n$ is a set of \emph{propositional variables} partitioned in $n$ disjoint subsets, one for each agent;
\item $S=2^\Phi$ is the set of \emph{states}, corresponding to all valuations over $\Phi$;

\item $d: N\times S \to (2^{\act} \setminus \emptyset)$, for $\act = 2^{\Phi}$, is the \emph{protocol function}, such that $d(i,s)\subseteq {\act_i}$ for $\act_i=2^{\Phi_i}$;

\item $\tau: S\times \act^n \to S$ is the \emph{transition function} such that $\tau(s, \alpha_1,\dots,\alpha_n) = \bigcup_{i \in N} \alpha_i $.
\end{itemize}
\end{definition}

Intuitively, a CGS-EPC describes the interactions of a group $N$ of
agents, each one of them controlling (exclusively) a set $\Phi_i
\subseteq \Phi$ of propositional atoms.  The state of the CGS is an
evaluation of the atoms in $\Phi$. In each such state the protocol
function returns which actions an agent can execute. 

The intuitive meaning of action $\alpha_i \in d(i,s)$ is ``assign
true to all atoms in $\alpha_i$, and false to all atoms in $\Phi_i
\setminus \alpha_i$''. The $idle_s$ action can be introduced as 
$\{ p \in \Phi_i \mid s(p) = 1 \}$, for every $i \in N$, $s \in S$.  
With an abuse of
notation we write $d(i,s) = \alpha$ whenever $d(i,s)$ is a singleton
$\{ \alpha \}$.

We equally see each state $s \in S$ as a function $s : \Phi
\to \{0,1\}$ returning the truth value of a propositional variable
in~$s$, so that $s(p) = 1$ iff $p \in s$.  Given $\alpha = (\alpha_1,
\dots, \alpha_n) \in \act^n$,
we equally see each $\alpha_i \subseteq \Phi_i$ as a function
$\alpha_i : \Phi_i \to \{0,1\}$
returning the choice of agent $i$ for $p$ under action $\alpha$.


We now introduce a generalisation of concurrent game structures for
propositional control. Namely, we relax the exclusivity requirement on
the control of propositional variables, thus introducing
concurrent game structures with shared propositional control CGS-SPC.
\begin{definition}[CGS-SPC]
A \emph{concurrent game structure with shared propositional control} is
  a tuple $\G = \langle N, \Phi_0, \dots, \Phi_n,$ $ S, d, \tau\rangle$ such that:

\begin{itemize}
\item $N$, $S$, and $d$ are defined as in Def.~\ref{def:CGS-PC} with $\act = 2^{\Phi \setminus \Phi_0}$;
\item $\Phi = \Phi_0\cup \Phi_1\cup \dots \cup \Phi_n$ is a set of
  \emph{propositional variables}, where $\Phi_0 \cup \Phi_1\cup \dots \cup
  \Phi_n$ is not necessarily a partition and $\Phi_0 = \Phi \setminus(\Phi_1\cup \dots \cup
  \Phi_n) $;
\item $\tau : S \times \act^n \to S$ is the \emph{transition function}.
\end{itemize}
\end{definition}

Observe that in CGS-SPC the same atom can be controlled by multiple agents, and propositional control is not exhaustive. 
Additionally, the actions in $\act$ do not take into account
propositional variables in $\Phi_0$ because they are not controlled by anyone (though their truth value might change according to the transition function). 
The transition function takes care of combining the various actions and producing a
consistent successor state according to some rule.  
Simple examples of such rules include introducing a threshold $m_p \in \mathbb{N}$ for
every variable $p$, thus setting $p \in \tau(s,\alpha)$ iff the number
of agents $i$ with $p \in \alpha_i$ is greater than $m_p$.
This generalises Gerbrandy's consensus games 
\cite{Gerbrandy06}.\footnote{The definition of $\tau$ as an arbitrary function might seem too general. Nonetheless, such a definition is needed to represent complex aggregation procedures such as those used in the games described in Sections~\ref{sec:influencegames} and \ref{sec:aggregationgames}.}

Clearly, CGS-EPC can be seen as a special case of CGS-SPC in which
every atom is controlled exactly by a single agent, and therefore
$\{\Phi_0, \dots, \Phi_n\}$ is a partition of $\Phi$. Moreover, $\tau$
is given in a specific form as per Definition \ref{def:CGS-PC}.  


\subsection{Logics for Time and Strategies}\label{sec:logics}

To express relevant properties of CGS, we present the Linear-time
Temporal Logic $\LTL$ \cite{P77} and the Alternating-time Temporal Logic
$\ATL^*$
\cite{AHK02}. 
Firstly, state formulas $\varphi$ and path formulas
$\psi$ in $\ATL^*$ are defined by the following BNF:
\begin{eqnarray*}
 \varphi & ::= & p \mid \neg\varphi \mid \varphi \lor \varphi \mid
 \atlop{C}\psi\\
 \psi & ::= & \varphi \mid \neg \psi \mid \psi \lor \psi  \mid \nextop{\psi} \mid \until{\psi}{\psi}
\end{eqnarray*}
where $p \in \Phi$ and $C \in 2^{N}$. The intuitive reading of $ \atlop{C}\psi$ is ``coalition $C$ has a strategy to enforce $\psi$'', that of $\nextop{\psi}$ is ``$\psi$ holds at the next state'' and that of $\until{\psi}{\varphi}$ is ``$\psi$ will hold until $\varphi$ holds''.

The BNF for the language of $\ATL$
consists of all state formulas where $\psi$ is either $\nextop{\varphi}$ or $\until{\varphi}{\varphi}$.  
On the other hand, the language of
$\LTL$ consists of all path formulas in $\ATL^*$, whose state formulas
are propositional atoms only. That is, formulas in $\LTL$ are defined
by the following BNF:
\begin{eqnarray*}
 \psi & ::= & p \mid \neg \psi \mid \psi \lor \psi \mid \nextop{\psi}
 \mid \until{\psi}{\psi}
\end{eqnarray*}

  Truth conditions of $\LTL$ and $\ATL^*$ formulas are defined
  with respect to concurrent game structures, such as the CGS-EPC and CGS-SPC
  introduced above. In order to do so, we first have to provide some
  additional notation.

The set of {\em enabled joint actions} at some state $s$ is defined as 
 $Act(s) = \{ \alpha \in \act^n \mid \alpha_i \in d(i, s) \text{ for every } i$ $\in N \}$.  
Then, the set of {\em successors} of $s$
is given as $Succ(s) = \{ \tau(s, \alpha) \mid \alpha \in Act(s) \}$.
Every $Succ(s)$ is non-empty because $d(i,s) \neq \emptyset$.
An infinite sequence of states $\lambda = s_0s_1\dots$ is a
\emph{computation} or a \emph{path} if $s_{k{+}1}\in Succ(s_k)$ for all $k \ge 0$. For
every computation $\lambda$ and $k \ge 0$, $\lambda[k, \infty] = s_k,
s_{k{+}1}, \dots$ denotes the suffix of $\lambda$ starting from $s_k$. 
Notice that $\lambda[k, \infty]$ is also a computation. 
When $\lambda$ is clear from the context, we denote with $\alpha[k]$ the action such that $\lambda[k{+}1]=\tau(\lambda[k], \alpha[k])$.

A \emph{memoryless strategy} for agent $i \in N$ is a function
 $\sigma_i : S \to \act_i$ such that $\sigma_i(s) \in d(i,s)$,
 returning an action for each state. 
 For simplicity, we will assume in the rest of the paper that agents have memoryless strategies.
 
 We let $\atlstr{C}$ be a {\em
   joint strategy} for coalition $C \subseteq N$, i.e., a function
 returning for each agent $i \in C$, the individual strategy
 $\sigma_i$. For notational convenience we write $\Str$ for
 $\Str_{N}$.  The set $\outset{s}{\Str_C}$ includes all computations
 $\lambda = s_0s_1\dots$ such that ($a$) $s_0 = s$; and ($b$) for all
 $k \ge 0$, there is $\alpha \in Act(s)$ such that $\Str_C(i)(s_k) =
 \alpha_i$ for all $i \in C$, and $\tau(s_k, \alpha) = s_{k{+}1}$. Observe
 that $\outset{s}{\Str}$ is a singleton.

We are now ready to define the truth conditions for $\LTL$ and
$\ATL^*$ formulas with respect to ~a CGS-SPC $\G$. Formulas in $\ATL^*$ are
interpreted on states, while formulas in $\LTL$ are
interpreted on computations.
\begin{center}
\begin{tabular}{lcl}
$(\G, s) \models p$ & iff & $s(p) = 1$\\ 
$(\G, s) \models \lnot \varphi$ & iff & $(\G, s) \not\models \varphi$ \\
$(\G, s) \models \varphi_1 \lor \varphi_2$ & iff & $(\G, s) \models \varphi_1 \text{ or } (\G, s) \models \varphi_2$\\ 
$(\G,s) \models \atlop{C}\psi$ & iff & for some $\Str_C$, for all $\lambda \in \outset{s}{\Str_C}$, $(\G, \lambda) \models \psi$\\
$(\G, \lambda) \models \varphi$ & iff & $(\G, \lambda[0]) \models \varphi$\\
$(\G, \lambda) \models \lnot \psi$ & iff & $(\G, \lambda) \not\models  \psi$\\
$(\G, \lambda) \models  \psi_1 \lor \psi_2$ & iff & $(\G, \lambda) \models  \psi_1 \text{ or } \G, \lambda \models  \psi_2$\\
$(\G, \lambda) \models  \nextop{\psi}$ & iff & $(\G, \lambda[1, \infty]) \models \psi$\\ 
$(\G, \lambda) \models \until{\psi_1}{\psi_2}$ & iff & for some  $i \ge 0$, $(\G, \lambda[i, \infty]) \models \psi_2$ and $(\G, \lambda[j, \infty]) \models \psi_1$ for all $0 \le j < i$
\end{tabular}
\end{center}

We define below the model checking problem for this context.
\begin{definition}[Model Checking Problem]\label{def:modelchecking}
Given a CGS-SPC $\G$, a state $s \in S$, and an $\ATL^*$-formula
$\varphi$, determine whether $(\G, s) \models \varphi$.
\end{definition}

It is well-known that model checking for $\ATL^*$ on general concurrent
game structures is 2EXPTIME-complete \cite{AHK02}. Belardinelli and Herzig proved that
model checking $\ATL$ on CGS-EPC is
$\Delta^P_3$-complete \cite{BH16}. Hereafter we consider the general case of
CGS-SPC and $\ATL^*$.


\section{Examples of Shared Control}\label{sec:games}


In this section we take three examples of iterated games from the
literature, namely \emph{iterated boolean
games} \cite{GutierrezEtAlIC20015}, \emph{influence
games} \cite{GLNP17}, and \emph{aggregation
games} \cite{GrandiEtAlIJCAI2015}, and we show that they are all
instances of our definition of a CGS-SPC.

\subsection{Iterated Boolean Games}

We make use of CGS-EPC to introduce iterated boolean games with $\LTL$
goals as studied by Gutierrez \emph{et al.} \cite{GHW13,GutierrezEtAlIC20015}.
An {\em iterated boolean game} is a tuple $\langle \G, \gamma_1,
\ldots, \gamma_n \rangle$ such that ($i$) $\G$ is a CGS-EPC with a
trivial protocol (i.e., for every $i \in N$, $s \in S$, $d(i,s) =
\act_i$); and ($ii$) for every $i \in N$, the {\em goal} $\gamma_i$
is an $\LTL$-formula.

We can generalise the above to {\em iterated boolean games with shared
control} as follows:
\begin{definition}
An \emph{iterated boolean game with shared control} is a tuple $\langle \G, \gamma_1, \ldots, \gamma_n \rangle$ such that 
\begin{itemize}
\item [(i)] $\G$ is a CGS-SPC;
\item [(ii)] for every $i \in N$, the {\em goal} $\gamma_i$ is an $\LTL$-formula.
\end{itemize}
\end{definition}
Observe that function $\tau$ is thus no longer trivial.
Just like CGS-SPC generalise CGS-EPC, iterated boolean games with
shared control generalise standard iterated boolean
games.
In particular, the existence of a winning strategy can be checked via the satisfaction of an  $\ATL^*$-formula:

\begin{proposition}
An agent $i$ in an iterated boolean game has a winning strategy for goal $\gamma_i$ and state $s$ if and only if formula $\atlop{\{ i \}}\gamma_i$ is satisfied in ($\G$,s). 
\end{proposition}


\begin{example}
Consider an iterated boolean game with shared control for agents $\{1,2\}$ and issues $\{p,q\}$, such that $\Phi_1 = \{p\}$ and $\Phi_2 =\{p,q\}$. Suppose that for all states $s$ the transition function is such that $\tau(s,\alpha)(q) = \alpha_2(q)$, being agent $2$ the only agent controlling $q$, while $\tau(s,\alpha)(p) = 1$ iff $\alpha_1(p) = \alpha_2(p) = 1$. We thus have that 
$(\G, s) \models \atlop{ \{1,2\} }\nextop p$ and 
$(\G, s) \models \lnot\atlop{ \{1\} }\nextop q$ for all $s$. 
\end{example}


\subsection{Influence Games}\label{sec:influencegames}

Influence games model strategic aspects of opinion diffusion on a
social network. They are based on a set of variables $\op{i}{p}$ for
"agent $i$ has the opinion $p$" and $\vis{i}{p}$ for "agent $i$ uses her influence power over $p$". Agents have binary opinions over all issues; hence $\lnot\op{i}{p}$ reads "agent $i$ has the opinion $\lnot p$".

Goals are expressed in $\LTL$ with propositional variables $\{\op{i}{p},$ $\vis{i}{p} \mid i\in N, p\in \Phi \}$. 
We define an influence game in a compact way below, pointing to the work of Grandi \emph{et al.} \cite{GLNP17} for more details.


\begin{definition}\label{def:influence}
An \emph{influence game} is a tuple $ IG= \langle N, \Phi, E,$ $ S_0,$ $\{F_{i,\textit{Inf(i)}}\}_{i \in N}, \{\gamma_i\}_{i\in N}\rangle$ where:
\begin{itemize}
\item $N = \{1, \dots, n\}$ is a set of \emph{agents};
\item $\Phi = \{1, \dots, m\}$ is a set of \emph{issues};
\item $E \subseteq N \times N$ is a directed irreflexive graph representing the \emph{influence network};
\item $S_0 \in \stateset$ is the \emph{initial state}, where states in $\stateset$ are tuples $(\B, \V)$, where $\B = (B_1, \dots, B_n)$ is a profile of \emph{private opinions} $B_i : \Phi \to \{0,1\}$ indicating the opinion of agent $i$ on variable $p$, and $\V = (V_1, \dots, V_n)$ is a profile of \emph{visibilities} $V_i : \Phi \to \{0,1\}$ indicating whether agent $i$ is using her influence power over $p$;
\item $F_{i, \textit{Inf(i)}}$ is the unanimous \emph{aggregation function} 
associating a new private opinion for agent $i$ based on agent i's current opinion and the visible opinions of $i$'s influencers in \textit{Inf(i)};
\item $\gamma_i$ is agent $i$'s \emph{individual goal}, i.e., an $\LTL$ formula.
\end{itemize}
\end{definition}

Influence games are repeated games in which individuals decide whether to disclose their opinions (i.e., use their influence power over issues) or not. Once the disclosure has taken place, opinions are updated by aggregating the visible opinions of the influencers of each agent (i.e., the nodes having an outgoing edge terminating in the agent's node).

We associate to $IG = \langle N, \Phi, E, S_0, \{F_{i,\textit{Inf(i)}}\}_{i \in N}, \{\gamma_i\}_{i\in N}\rangle$ a CGS-SPC 
$\G' =  \langle N', \Phi'_0, \dots, \Phi'_n, S', d', \tau' \rangle$ 
by letting 
$N' = N$; 
$\Phi'_0 = \{\op{i}{p}\mid i \in \N, p \in \Phi\}$; 
$\Phi_i' = \{ \vis{i}{p} \mid p \in \Phi\}$ for $i \in N'$; 
$S' = 2^{\Phi'}$; 
$d'(i, s') = 2^{\Phi_i'}$ for $s' \in S'$; 
and finally for state $s'\in S'$ and action $\alpha'$ we let:
\[
\tau'(s', \alpha')(\varphi) \; = \;
\left\{
	\begin{array}{ll}
		\alpha'_i(\vis{i}{p}) & \mbox{if } \varphi = \vis{i}{p} \\
		F_{i,\textit{Inf(i)}}(\vec{a}, \vec{b})_{|p} & \mbox{if }  \varphi = \op{i}{p}
	\end{array}
\right.
\]
where vectors $\vec{a} = (a_1, \dots, a_{|\Phi|})$ and $\vec{b} = (b_1, \dots, b_{|\Phi|})$ are defined as follows, for $k \in \textit{Inf(i)}$: 
\begin{align*}
a_p &=  \begin{cases}
		1 & \mbox{if } \op{i}{p} \in s' \\
		0 & \mbox{otherwise }
	\end{cases}
\\
b_p & = \begin{cases}
		1 & \mbox{if } \alpha_k(\vis{k}{p}) = 1 \text{ and } \op{k}{p} \in s' \\
		0 & \mbox{if } \alpha_k(\vis{k}{p}) = 1 \text{ and } \op{k}{p} \not\in s'  \\
		? & \mbox{if } \alpha_k(\vis{k}{p}) = 0
	\end{cases}
\end{align*}
Vector $\vec{a}$ represents the opinion of agent $i$ over the issues at state $s'$, while vector $\vec{b}$ represents the opinions of $i$'s influencers over the issues, in case they are using their influencing. In particular, `?' indicates that the influencers of $i$ in \textit{Inf(i)} are not using their influence power.

\begin{proposition}\label{prop:infgam}
Agent $i$ in influence game IG has a winning strategy for goal $\gamma_i$ and state $S_0$ if and only if formula $\atlop{\{ i \}}\gamma_i$ is satisfied in the associated CGS-SPC and state $s'$ corresponding to $S_0$.
\end{proposition}

\begin{proof}[proof sketch]
Let $IG$ be an influence game and let $\G'$ be the CGS-SPC associated to it. 
Consider now an arbitrary agent $i$ and suppose that $i$ has a winning strategy in $IG$ for her goal $\gamma_i$ in $S_0$. 
A memoryless strategy $\sigma_i$ for agent $i$ in an influence game maps to each state actions of type $(\mathsf{reveal}(J), \mathsf{hide}(J'))$, where $J, J' \subseteq \Phi$ and $J \cap J' = \emptyset$. For any state $s$ in $IG$, consisting of a valuation of opinions and visibilities, consider the state $s'$ in $\G'$ where $B_i(p) = 1$ iff $\op{i}{p} \in s'$ and $V_i(p) = 1$ iff $\vis{i}{p} \in s'$. We now construct the following strategy for $\G'$:
\[
\sigma_i'(s') = \{\vis{i}{p} \mid p \in J \text{ for } \sigma_i(s) = (\mathsf{reveal}(J), \mathsf{hide}(J'))\}
\]
By the semantics of the $\atlop{\{i\}}$ operator provided in Section \ref{sec:logics}, and by the standard game-theoretic definition of winning strategy, the statement follows easily from our construction of $\G'$.
%
\end{proof}

The above translation allowed to shed light over the control structure of the variables of type $\op{i}{p}$. In fact, we can now see that $\op{i}{p} \in \Phi'_0$ for all $i \in N$ and $p \in \Phi$.


\subsection{Aggregation Games}\label{sec:aggregationgames}

Individuals facing a collective decision, such as members of a hiring committee or a parliamentary body, are provided with individual goals specified on the outcome of the voting process --- outcome that is jointly controlled by all individuals in the group. For instance, a vote on a single binary issue using the majority rule corresponds to a game with one single variable controlled by all individuals, the majority rule playing the role of the transition function.

Similar situations have been modelled as one-shot games called \emph{aggregation games} \cite{GrandiEtAlIJCAI2015}, and we now extend this definition to the case of iterated decisions:
%

\begin{definition}\label{def:aggregation}
An \emph{iterated aggregation game} is a tuple $AG=\langle N, \Phi, F, \gamma_1, \dots, \gamma_n\rangle$ such that:

\begin{itemize}
\item N is a set of \emph{agents};
\item $\Phi =  \{p_1, \dots, p_m\}$  are variables representing \emph{issues};
\item $F : \{0,1\}^{N \times \Phi} \to \{0,1\}$ is an \emph{aggregation function}, that is, 
a boolean function associating a collective decision with the individual opinion of the agents on the issues;
\item $\gamma_i$ for $i \in N$ is an \emph{individual goal} for each agent, that is, 
a formula in the $\LTL$ language constructed over $\Phi$.
\end{itemize}

\end{definition}

Individuals at each stage of an aggregation game only have information about the current valuation of variables in $\Phi$, resulting from the aggregation of their individual opinions. 
Analogously to Proposition \ref{prop:infgam}, we can obtain the following result:

\begin{proposition}
An iterated aggregation game $AG$ is an instance of a CGS-SPC. 
More precisely, agent $i$ in $AG$ has a winning strategy for goal $\gamma_i$ in $s$ if and only if formula $\atlop{\{i\}}\gamma_i$ is satisfied in the associated CGS-SPC in the corresponding state $s'$.
\end{proposition}

\begin{proof}[proof sketch]
Starting from an iterated aggregation game $AG=\langle N, \Phi,$ $ F, \gamma_1, \dots, \gamma_n\rangle$, construct a CGS-SPC $\G' =  \langle N', \Phi', S', d', \tau' \rangle$ as follows. 
Let $N'=N$; 
$\Phi'_i=\Phi$ for all $i=1,\dots,n$; and 
$\Phi'_0=\emptyset$. 
Hence, each agent controls all variables.
Let the set of actions available to each player be $d'(i,s)=2^{\Phi'}$ for all $i$ and $s$, and the transition function $\tau'$ be such that $\tau'(s,\alpha_1,\dots,\alpha_n)=F(\alpha_1,\dots,\alpha_n)$. The statement then follows easily.
\end{proof}

A notable example of an iterated aggregation game is the setting of iterative voting (see, e.g., \cite{MeirEtAl2010,LevRosenscheinAAMAS2012,ObraztsovaEtAlAAAI2015}). 
In this setting, individuals hold preferences about a set of candidates and 
iteratively manipulate the result of the election in their favour until a converging state is reached.
Similar situations can easily be modelled as iterated aggregation games, which have the advantage of allowing for a more refined specification of preferences via the use of more complex goals.


\section{Restoring Exclusive Control}\label{main_result}
 

In this section we prove the main result of the paper, namely 
that the shared control of a CGS-SPC can be simulated in a CGS-EPC having exclusive control.  
In particular, any
specification in $\ATL^*$ satisfied in some CGS-SPC can be translated
in polynomial time into an $\ATL^*$-formula satisfied in a CGS-EPC.
To do so, we introduce a dummy agent to simulate the aggregation
function. 
Moreover, we make use of an additional `turn-taking' atom which allows us to distinguish
the states where the agents choose their actions from those in which the
aggregation process takes place.

We begin by inductively defining a translation function $tr$ within $\ATL^*$. 
Intuitively, $tr$ translates every $\ATL^*$-formula $\chi$ into a formula $tr(\chi)$ having roughly the
same meaning, but where the one-step `next' operator is replaced by two `next' steps:

\begin{center}
\begin{tabular}{lcl}
$tr(p)$& $=$ & $p$ \\
$tr(\lnot \chi)$ & $=$ & $\lnot tr(\chi)$  \\
$tr(\chi \lor \chi')$ & $=$ & $tr(\chi) \lor tr(\chi')$ \\
$tr(\nextop \chi)$ & $=$ & $\nextop\nextop tr(\chi)$ \\
$tr(\until{\chi}{\chi'})$ & $=$& $\until{tr(\chi)}{tr(\chi')}$ \\
$tr(\atlop{C} \chi)$ & $=$ & $\atlop{C} tr(\chi)$ 
\end{tabular}
\end{center}
where $p \in \Phi$, $C \subseteq N$, and $\chi$, $\chi'$ are either
state- or path-formulas as suitable.   Clearly, the translation is polynomial.

We then map a given CGS-SPC to a CGS-EPC.

\begin{definition}\label{def:associatedgame}
Let $\G = \langle N, \Phi_0,\dots,\Phi_n, S, d, \tau\rangle$ be a CGS-SPC.
The CGS-EPC corresponding to $\G$ is 
$\G'=\langle N', \Phi'_1,\dots,$ $\Phi'_n, S', d', \tau' \rangle$ where:
\begin{itemize}
\item 
$N' = N \cup \{*\}$;
\item 
$\Phi' = \Phi \cup \{turn\} \cup \{c_{ip} \mid i \in N \text{ and } p \in \Phi_i\}$ 
and $\Phi'$ is partitioned as follows, for agents in $N'$:
	\begin{align*}
	\Phi'_i & = \{ c_{ip} \in \Phi' \mid p \in \Phi_i\} \\
	\Phi'_* & = \{ turn\} \cup \Phi
	\end{align*}
\item 
$S' = 2^{\Phi'}$. For every $s' \in S'$, let 
$s =  (s' \cap \Phi) \in S$ be the restriction of $s'$ on $\Phi$;
\item 
$d'$ is defined according to the truth value of $turn$ in $s'$. 
Specifically, given $\alpha_i \in \act_i$, let $\alpha'_i = \{ c_{ip} \in \Phi'_i \mid p \in \alpha_i \} \in \act'_i$. Then, for $i\in N$ we let:
$$d'(i,s') = 
\begin{cases}
\{ \alpha'_i \in \act'_i \mid \alpha_i \in d(i,s) \} & \mbox{if }  s'(turn) = 0\\
\emptyset &\mbox{if } s'(turn) = 1
\end{cases}
$$
For agent * we define:
$$d'(*,s') = 
\begin{cases}
+turn    & \mbox{if }  s'(turn) = 0\\
\tau(s, \alpha),  \text{ for } \alpha_i(p) = s'(c_{ip}) &\mbox{if } s'(turn) = 1
\end{cases}
$$
where $+turn = idle_s \cup \{ turn \}$. 

%


\item $\tau'$ is defined as per Def.~\ref{def:CGS-PC}, that is,
  $\tau'(s', \alpha') = \bigcup_{i \in N'} \alpha'_i$.
\end{itemize}
\end{definition}

Intuitively, in the CGS-EPC $\G'$ every agent $i \in N$ manipulates
local copies $c_{ip}$ of atoms $p \in \Phi$. The aggregation function
$\tau$ in $\G$ is mimicked by the dummy agent $*$, whose role is to
observe the values of the various $c_{ip}$, then perform an action to
aggregate them and set the value of $p$ accordingly.
Observe that agent $*$ acts only when the $turn$ variable is true, in which case all the
other agents set all their variables to false, i.e., they all play $\emptyset$.
This is to ensure the correspondence between memory-less strategies of $\G$ and $\G'$, as shown in Lemma~\ref{aux}.

Note also that the size of game $\G'$ is polynomial in the size of $\G$, and that $\G'$ can be constructed in polynomial time from $\G$. To see this, observe that an upper bound on the number of variables is $\N \times \Phi$.

%
Recall that we can associate to each state $s'\in S'$ a state $s=s' \cap \Phi$ in $S$. 
For the other direction, given a state $s\in S$, there are multiple states $s'$ that agree with $s$ on $\Phi$. The purpose of the next definition is to designate one such state as the canonical one.

\begin{definition}\label{def:canonicalstate}
For every $s \in S$, we define the \emph{canonical state} $s'_\star=\{s'\in S' \mid s' \cap \Phi = s \text{ and } s(p)=0 \text{ for } p\not\in \Phi\}$.
\end{definition}

\noindent
Observe that, in particular, in all canonical states atom $turn$ is false. As an example, consider $\Phi = \{p,q\}$ and $N = \{1,2\}$. Let then $\Phi_1 = \{p\}$ and $\Phi_2 = \{p,q\}$.We thus have that $\Phi' = \{p,q,c_{1p},c_{2p},c_{2q},turn\}$. If $s = \{p\}$, we have for instance that $s' \cap \Phi = s$ for $s' = \{p, c_{1p}\}$. On the other hand, $s'_\star = \{p\}$.

%
%
%
We now move to define a correspondence between paths of $\G$ and $\G'$.
For notational convenience, we indicate with $\lambda[k]_{|\Phi}=\lambda[k] \cap \Phi$, the restriction of state $\lambda[k]$ to variables in $\Phi$.
Given a path $\lambda'$ of $\G'$, consider the unique infinite sequence of states $\lambda$ associated to $\lambda'$ defined as follows:
$$\lambda[k] = \lambda'[2k]_{|\Phi} =
\lambda'[2k{+}1]_{|\Phi} \text{ for all } k\in
\mathbb{N}. \;\;\; (\dagger)$$
On the other hand, there are multiple sequences $\lambda'$ that can be associated with a path $\lambda$, so that $(\dagger)$ holds true. In fact, we only know how the variables in $\Phi$ behave, while the truth values of the other variables can vary. We now make use of condition $(\dagger)$ to characterise the paths of $\G$ and $\G'$ that can be associated:


\begin{lemma} \label{aux11}
Given a CGS-SPC $ \G$ and the corresponding CGS-EPC $\G'$, the following is the case: 

\begin{enumerate}
\item for all paths $\lambda'$ of $\G'$, sequence $\lambda$ satisfying condition $(\dagger)$ is a path of $\G$; 

\item for all paths $\lambda$ of $\G$, for all sequences $\lambda'$ satisfying $(\dagger)$, $\lambda'$ is a path of $\G'$ iff
for all $k$ there exists a $\G$-action $\alpha[k]$ such that $\lambda[k] \xrightarrow{\alpha[k]} \lambda[k{+}1]$ and states $\lambda'[2k{+}1]$ and $\lambda'[2k{+}2]$ can be obtained from state $\lambda'[2k]$ by performing actions $(\alpha'_1, \ldots, \alpha'_n,$ $+turn)$ and then $(\emptyset_1, \ldots, \emptyset_n,$ $\tau(\lambda'[2k{+}1]_{|\Phi}, \alpha))$.
\end{enumerate}
\end{lemma}

\begin{proof}
We first prove $(1)$ by showing that
$\lambda$ is a path of $\G$, i.e., that for every $k$ there is an action $\alpha$ that leads from $\lambda[k]$ to $\lambda[k{+}1]$. 
Suppose that $\lambda'[2k] \xrightarrow{\alpha'[2k]}
  \lambda'[2k{+}1] \xrightarrow{\alpha'[2k{+}1]} \lambda'[2(k{+}1)]$ for action
  $\alpha'[2k] = (\alpha'_1, \ldots, \alpha'_n, +turn)$ and action
  $\alpha'[2k{+}1] = (\emptyset_1, \ldots, \emptyset_n, $$
  \tau(\lambda'[2k{+}1]_{|\Phi}, \alpha))$.  
Then, we observe that we can move from state
$\lambda[k] = \lambda'[2k]_{|\Phi} =
  \lambda'[2k{+}1]_{|\Phi}$ to $\lambda[k{+}1] = \lambda'[2k{+}2]_{|\Phi}$ by performing action $(\alpha_1, \ldots,
  \alpha_n)$ such that $\alpha_i = \{p \in \Phi
  \mid c_{ip} \in \alpha'_i \}$ for every $i \in N$.  

As for (2), the right-to-left direction is clear.
For the left-to-right direction, let $\lambda'$ be a path associated to $\lambda$. From $(\dagger)$ we know that for any $k$ we have that $\lambda'[2k]_{|\Phi}=\lambda[k]$ and $\lambda'[2k{+}2]_{|\Phi}=\lambda[k{+}1]$.
Now by Definition~\ref{def:associatedgame}, the only actions available to the players at $\lambda'[2k]$ are of the form $(\alpha'_1, \ldots, \alpha'_n, +turn)$, and the only action available at $\lambda'[2k{+}1]$ is $(\emptyset_1, \ldots, \emptyset_n, \tau(\lambda'[2k{+}1]_{|\Phi}, \alpha))$. 
We can thus obtain the desired result by considering action $\alpha[k]=(\alpha_1, \ldots, \alpha_n)$, where $\alpha_i=\{p\in\Phi_i \mid c_{ip}\in\alpha_i'\}$ for each $i \in N$, and by observing that by $(\dagger)$ we have $\tau(\lambda'[2k+1]_{|\Phi}, \alpha) = \tau(\lambda[k], \alpha)$.
\end{proof}

{\footnotesize
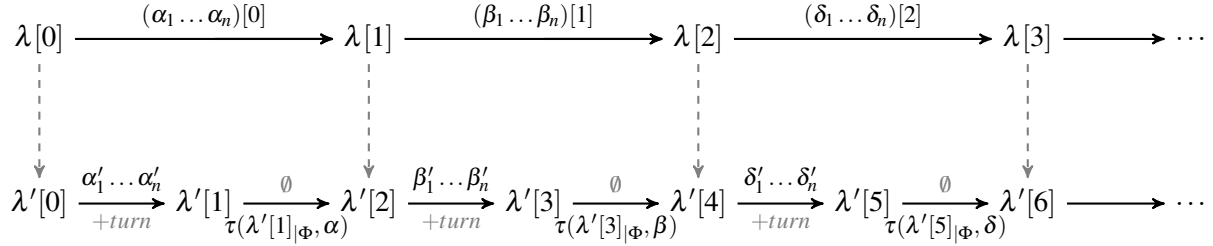
\begin{figure*}
\begin{tikzpicture}[->,>=stealth',shorten >=0pt,auto,node distance=2.19cm,
                    thick,main node/.style={font=\sffamily}]

  \node[main node] (3)  {$\lambda[0]$};
  \node[main node] (4) [below of=3] {$\lambda'[0]$};
  \node[main node] (5) [right of=3] {};
  \node[main node] (6) [below of=5] {$\lambda'[1]$};
\node[main node] (7) [right of=5] {$\lambda[1]$};
\node[main node] (8) [right of=6] {$\lambda'[2]$};
\node[main node] (9) [right of=7] {};
\node[main node] (10) [right of=8] {$\lambda'[3]$};
\node[main node] (11) [right of=9] {$\lambda[2]$};
\node[main node] (12) [right of=10] {$\lambda'[4]$};
\node[main node] (13) [right of=11] {};
\node[main node] (14) [right of=12] {$\lambda'[5]$};
\node[main node] (15) [right of=13] {$\lambda[3]$};
\node[main node] (16) [right of=14] {$\lambda'[6]$};
\node[main node] (17) [right of=15] {$\dots$};
\node[main node] (18) [right of=16] {$\dots$};

  \path[every node/.style={font=\sffamily\footnotesize}]
    (3) edge node {$(\alpha_1 \dots \alpha_n)[0]$} (7)
        edge[gray, dashed] node {} (4)
    (4) edge node[above] {$\alpha'_1\dots\alpha'_n$} node[gray, below] {$+ turn$} (6)
    (5) edge node {} (7)
     (6) edge node[gray, above] {$\emptyset$} node[below] {$\tau(\lambda'[1]_{|\Phi}, \alpha)$} (8)
    (7) edge node {$(\beta_1 \dots \beta_n)[1]$} (11)
    edge[gray, dashed] node {} (8)
    (8) edge node[above] {$\beta'_1\dots\beta'_n $} node[gray, below] {$+ turn$} (10)       
    (9) edge node {} (11)
    (10) edge node[gray,above] {$\emptyset$} node[below] {$\tau(\lambda'[3]_{|\Phi}, \beta)$}  (12)  
      (11) edge node {$(\delta_1 \dots \delta_n)[2]$} (15)
   edge[gray, dashed] node {} (12)
    (12) edge node[above] {$\delta'_1\dots\delta'_n$} node[below,gray] {$+ turn$} (14) 
    (14) edge node[above,gray] {$\emptyset$} node[below] {$\tau(\lambda'[5]_{|\Phi}, \delta)$}  (16) 
    (15) edge node {} (17) 
    edge[gray, dashed] node {} (16)
    (16) edge node {} (18) 
    ;
\end{tikzpicture}

\caption{A path $\lambda$ in a CGS-SPC $\G$ and its associated path  $\lambda'$ in a CGS-EPC $\G'$.  
\label{fig1}}
\end{figure*}
}

Figure 1 illustrates the construction of the two paths $\lambda$ and $\lambda'$ in the proof of Lemma \ref{aux11}. 
In particular, the second part of the lemma characterises the set of $\G'$-paths $\lambda'$ associated to a $\G$-path $\lambda$: for any sequence of $\G$-actions that 
can generate path $\lambda$, we can construct a distinct $\G'$-path $\lambda'$ that corresponds to $\lambda$, where the sequence of actions can be reconstructed by 
reading the values of the variables in $\Phi_i'$ in odd states $\lambda[2k+1]$.

From this set of $\G'$-paths $\lambda'$ we can specify a subset of \emph{canonical} paths as follows:

\begin{definition}\label{def:canonicalpath}
For a path $\lambda$ of $\G$, a \emph{canonical} associated path $\lambda'_\star$ of $\G'$ is any path $\lambda'$ such that $(\dagger)$ holds and $\lambda'[0]=\lambda[0]'_\star$. 
\end{definition}

\noindent
That is, a canonical path $\lambda'$ associated to $\lambda$ starts from the canonical state $\lambda[0]'_\star$ associated to $\lambda[0]$. The following example clarifies the concepts just introduced.

\begin{example}
Consider a CGS-SPC $\G$ with $N = \{1,2\}$ and $\Phi = \{p,q\}$ such that $\Phi_1 = \{p\}$ and $\Phi_2 = \{p,q\}$. Let $d(i,s) = 2^{\Phi_i}$ for all $i \in N$ and $s \in S$, and let $\tau(s, \alpha)(p) = 0$ if and only if $\alpha_1(p) = \alpha_2(p) = 0$, while $\tau(s, \alpha)(q) = \alpha_2(q)$ for all $s \in S$. Namely, issue $p$ becomes true if at least one agent makes it true, while issue $q$ follows the decision of agent 2. Let now $\lambda = s_0s_1\dots$ be a path of $\G$ such that $s_0 = \{p\}$ and $s_1 = \{p,q\}$. Observe that there are multiple actions $\alpha$ such that $\tau(s_0, \alpha) = s_1$: namely, the one where both agents set $p$ to true, or where just one of them does (and agent 2 sets $q$ to true).

Construct now the CGS-EPC $\G'$ as in Definition \ref{def:associatedgame} and consider the following four sequences $\lambda' = s'_0s'_1s'_2\dots$ where:
\begin{itemize}
\item[(a)] $s'_0 = \{p\}$, $s'_1 = \{c_{1p}, c_{2p}, c_{2q}, p, turn\}$, $s'_2 = \{p,q\}$, $\dots$
\item[(b)] $s'_0 = \{p\}$, $s'_1 = \{c_{1p}, c_{2q}, p, turn\}$, $s'_2 = \{p,q\}$, $\dots$
\item[(c)] $s'_0 = \{p, c_{1p}\}$, $s'_1 = \{c_{1p}, c_{2q}, p, turn\}$, $s'_2 = \{p,q\}$, $\dots$
\item[(d)] $s'_0 = \{p\}$, $s'_1 = \{c_{2q}, p, turn\}$, $s'_2=\{p,q\}$, $\dots$
\end{itemize}
Observe that (a) and (b) are both examples of canonical paths (up to the considered state), corresponding to two actions that might have led from $s_0$ to $s_1$ in$\G$. On the other hand, (c) is a non-example while being a path of $\G'$ satisfying $(\dagger)$, since $s'_0$ is not canonical. Finally, sequence (d) satisfies $(\dagger)$ but it is not a path of $\G'$, since it is not possible to obtain $s'_2$ from $s'_1$.
\end{example}


The next result extends the statement of Lemma~\ref{aux11} to paths
generated by a specific strategy.
Given a $\G'$-strategy $\Str'_C$ and a state $s'\in S'$, let $\Pi(\outset{s'}{\Str'_C})=\{\lambda  \mid \lambda' \in \outset{s'}{\Str'_C} \}$, i.e., all the ``projections'' of paths $\lambda'$ in $\outset{s'}{\Str'_C}$ to paths $\lambda$ in $\G$, obtained through $(\dagger)$.

\begin{lemma} \label{aux}
Given a CGS-SPC $ \G$, the corresponding CGS-EPC $ \G'$ is such that:
\begin{enumerate}
\item \label{aux1} for every joint strategy $\Str_C$ in $\G$, there exists a
  strategy $\Str'_C$ in $\G'$ such that for every state $s\in S$ we have that 
  $\Pi(\outset{s'_\star}{\Str'_C})=\outset{s}{\Str_C}$;
\item \label{aux2} for every joint strategy $\Str'_C$ in $\G'$, there exists a
  strategy $\Str_C$ in $\G$ such that for all \emph{canonical} states $s' \in S'$ we have that
  $\Pi(\outset{s'}{\Str'_C})=\outset{{s'}_{|\Phi}}{\Str_C}$.
\end{enumerate}
\end{lemma}


\begin{proof}[proof sketch]
We first prove (\ref{aux1}). Given strategy $\Str_C$ in $\G$, for $i
\in C$ define $\sigma_i'$ as follows:
\begin{eqnarray*}
\sigma_i'(s') & = &
	\begin{cases}
		\{c_{ip} \mid p \in \sigma_i(s) \text{ and } s={s'}_{|\Phi} \} & \mbox{if } s'(turn) = 0 \\
		\emptyset & \mbox{otherwise}  
	\end{cases}
\end{eqnarray*}
Observe that if $s'(turn) = 1$ agents in $C$ are obliged to play action $\emptyset$ by Definition \ref{def:associatedgame}, since it is their only available action.
By combining all definitions above, we get that $\Pi(\outset{s'_\star}{\Str'_C}) =\outset{s}{\Str_C}$ for an arbitrary state $s \in S$.

To prove (\ref{aux2}), we start from a strategy $\Str_C'$ in $\G'$. 
For any state $s\in S$, define $\sigma_i(s)=\{p\in\Phi_i \mid c_{ip}\in \sigma_i'(s'_\star)\}$.
Note that the assumption in Definition~\ref{def:associatedgame} that all variables outside of $\Phi$ are put to false at stage $2k{+}1$ in $\G'$ is crucial here.
In fact, without this assumption we would only be able to prove that $\Pi(\outset{s'}{\Str'_C})\supseteq \outset{{s'}_{|\Phi}}{\Str_C}$, as a strategy $\Str_C'$ may associate a different action to states $s_1'$ and $s_2'$ that coincide on $\Phi$ and that are realised in a path $\lambda'\in \outset{s'}{\Str'_C}$.
\end{proof}


By means of Lemma~\ref{aux} we are able to prove the main result of
this section.
\begin{thm}\label{thm:reduction}
Given any CGS-SPC $ \G$, the corresponding CGS-EPC $ \G'$ is such that
for all state-formulas $\varphi$ and path-formulas $\psi$ in $\ATL^*$ the following holds:
\begin{eqnarray*}
\text{for all } s\in S \;\;\; (\G, s) \models \varphi & \text{iff} & (\G', s'_\star) \models tr(\varphi)\\
\text{for all } \lambda \text{ of } \G \;\;\; (\G, \lambda) \models \psi & \text{iff} &  (\G', \lambda'_\star) \models tr(\psi)  \text{ for any } \lambda_\star'.
\end{eqnarray*}
\end{thm}

\begin{proof}
The proof is by induction on the structure of formulas $\varphi$ and
$\psi$.  The base case for $\varphi = p$ follows from the fact that $s
= {s'}_{|\Phi}$ for all $s'$ associated to $s$, and in particular also for $s'_\star$.
%
As to the inductive cases 
for boolean connectives, these follow immediately by the induction
hypothesis.

Now suppose that $\varphi = \atlop{C} \psi$. As to the left-to-right direction, assume that $(\G, s) \models \varphi$.
By the definition of the semantics, for some strategy $\Str_C$, for
all $\lambda \in \outset{s}{\Str_C}$, $(\G, \lambda) \models \psi$. 
By Lemma~\ref{aux}.\ref{aux1} we can find a strategy
$\Str'_C$ in $\G'$ such that 
$\Pi(\outset{s'_\star}{\Str'_C})=\outset{s}{\Str_C}$.
By induction hypothesis, we know that for all $\lambda \in \outset{s}{\Str_C}$ we have that $(\G', \lambda'_\star) \models tr(\psi)$.
These two facts combined imply that for all $\lambda' \in \outset{s'_\star}{\Str'_C}$ we have that $(\G', \lambda'_\star) \models tr(\psi)$, i.e., by the semantics, that $(\G', s'_\star) \models \atlop{C} tr(\psi)$, obtaining the desired result.
The right-to-left direction can be proved similarly, by using
Lemma~\ref{aux}.\ref{aux2}.

Further, if $\varphi$ is a state formula, $(\G, \lambda) \models \varphi$ iff $(\G, \lambda[0]) \models
\varphi$, iff by induction hypothesis $(\G', \lambda[0]'_\star) \models
tr(\varphi)$, that is, $(\G', \lambda'_\star) \models tr(\varphi)$.

For $\psi = \nextop \psi_1$, suppose that $(\G, \lambda[1, \infty])
\models \psi_1$. By induction hypothesis, this is the case if and only if $(\G',
(\lambda[1, \infty])'_\star) \models tr(\psi_1)$.
Recall that by $(\dagger)$, we have that $(\lambda[1, \infty])'_\star=\lambda'_\star[2, \infty]$. 
This is the case because, when moving from $\lambda$ to $\lambda'_\star$, we include an additional state $\lambda'_\star[1]$ in which the aggregation takes place.
Therefore, $(\G', \lambda'_\star[2, \infty]) \models
tr(\psi_1)$, that is, $(\G', \lambda'_\star) \models \nextop \nextop
tr(\psi_1) = tr(\psi)$.
The case for $\psi = \until{\psi_1}{\psi_2}$ is proved similarly.
\end{proof}

As a consequence of Theorem~\ref{thm:reduction}, if we want to model-check an $\ATL^*$-formula $\varphi$ at a state $s$ of an CGS-SPC $\G$, we can
check its translation $tr(\varphi)$ at the related state $s'_\star$ of the associated
CGS-EPC $\G'$. Together with the observation that both the associated game $\G'$ and the translation $\varphi$ are polynomial in the size of $\G$ and $\varphi$, we obtain the following:

\begin{cor}
The $\ATL^*$ model-checking problem for CGS-SPC can be reduced to the $\ATL^*$ model-checking problem for CGS-EPC.
\end{cor}

\section{Computational Complexity of Shared Control Structures} \label{applications}


The results proved in the previous sections allow us to obtain
complexity results for the model checking of an $\ATL^*$ (or $\ATL$)
specification $\varphi$ on a {\em pointed} CGS-SPC $(\G, s)$ defined in Definition \ref{def:modelchecking}.
\begin{thm}\label{thm:complexity}
The model-checking problem of $\ATL$ specifications in CGS-SPC is $\Delta_3^p$-complete.
\end{thm}

\begin{proof}
As for membership, given a pointed CGS-SPC $(\G, s)$ and an $\ATL$
specification $\varphi$, by the translation $tr$ introduced in
Section~\ref{main_result} and Theorem~\ref{thm:reduction} we have that
$(\G, s) \models \varphi\;\; \text{iff}\;\; (\G', s') \models
tr(\varphi)$. Also, we observe that the CGS-EPC $\G'$ is of size
polynomial in the size of $\G$, and that model checking $\ATL$
with respect to~CGS-EPC is $\Delta_3^p$-complete \cite{BH16}. For hardness,
it is sufficient to observe that CGS-EPC are a subclass of CGS-SPC.
\end{proof}

As for the verification of $\ATL^*$, we can immediately prove the
following result:

\begin{thm}\label{thm:complexity*}
The model-checking problem of $\ATL^*$ specifications in CGS-SPC is
PSPACE-complete.
\end{thm}
\begin{proof}
Membership follows by the PSPACE-algorithm for $\ATL^*$ on
general CGS \cite{BullingEtAl2010}. As for hardness, we observe that satisfiability of an
$\LTL$ formula $\varphi$ can be reduced to the model checking of the $\ATL^*$
formula $\atlop{1} \varphi$ on a CGS-SPC with a unique agent $1$.
\end{proof}


In Section~\ref{sec:games} we showed how three examples of iterated
games from the literature on strategic reasoning can be modelled as
CGS-SPC, and how the problem of determining the existence of a winning
strategy can therefore be reduced to model checking an
$\ATL^*$ specification.  Let $\textsc{E-WIN}(G,i)$ be the decision
problem of deciding whether agent $i$ has a memory-less winning
strategy in game $G$.  As an immediate consequences of
Theorem~\ref{thm:complexity*} we obtain:
\begin{cor}
If $G$ is an iterated boolean game with shared control,
$\textsc{E-WIN}(G,i)$ is in PSPACE.
\end{cor}

An analogous result cannot be obtained for influence and aggregation
games directly. Decision problems in these structures are typically
evaluated with respect to the number of agents and issues, and the size of the
CGS-SPCs associated to these games are already exponential in these
parameters.  Therefore, in line with previous results obtained in the
literature \cite{GLNP17}, we can only show the following:

\begin{cor}\label{complexity_influence}
If $G$ is an influence game or an aggregation game, then
$\textsc{E-WIN}(G,i)$ is in $PSPACE$ in the size of the associated
CGS-SPC.
\end{cor}

%


%
%


\section{Conclusion}\label{sec:conclusions}


In this contribution we have introduced a class of concurrent game
structures with shared propositional control, or CGS-SPC. Then, we
have interpreted popular logics for strategic reasoning $\ATL$ and
$\ATL^*$ on these structures. Most importantly, we have shown that
CGS-SPC are a general framework, whereby we can capture iterated
boolean games and their generalisation to shared control, as well as
influence and aggregation games. The main result of the paper shows
that the model checking problem for CGS-SPC can be reduced to the
verification of standard CGS with exclusive control, which in turn allows
us to establish a number of complexity results.

The results proved here open up several research directions.  Firstly,
in this paper we have focussed on the verification problem, but what
about satisfiability and validity?  The undecidability result provided
by Gerbrandy \cite{Gerbrandy06} for $\clpclogic$ with shared control does not
immediately transfer to CGS-SPC, as the relevant languages are
different: $\clpclogic$ includes normal modal `diamond-operators'
$\langle C \rangle$ and `box-operators' $[C] $, while our $ \atlop{C}
$ is non-normal.\footnote{We observe that, on the other hand,
following van der Hoek and Wooldridge \cite{CLPC2015} the fragment of the language of $\ATL$
without `until' can be embedded into that of $\clpclogic$ by
identifying $ \atlop{C}\nextop{\phi} $ with $\langle C \rangle
[N \setminus C] \phi$.  }

Further, given our reduction of CGS-SPC to CGS with exclusive control,
one may wonder what the benefits of our move to shared control are.
As our three examples have demonstrated, shared control allows
to model in a natural way complex interactions between agents
concerning the assignment of truth values to propositional variables.
The strategic aspects of these games remain largely unexplored, and clean 
characterisations of equilibria and other game-theoretic concepts seem rather hard to prove, 
supporting the use of automated verification in these context.
%

Compact representations of CGS with exclusive control are a thriving subject of research in the formal verification community (see, e.g., \cite{HoekLW06,JamrogaA07,HuangCS15}).
There, so-called reactive modules define for every action whether it is available by means of a boolean formula.
In future work we plan to investigate such compact representations for CGS with shared control.
This requires in particular a compact representation of the transition function $\tau$, which becomes more involved in the shared control setting.


Finally, we conclude by remarking that a key assumption on our CGS
(both with exclusive and shared control) is that agents have perfect
knowledge of the environment they are interacting in and with.
Indeed, in Definition \ref{def:associatedgame} the dummy agent $*$ is
able to mimick the aggregation function $\tau$ as she can observe the
values of $c_{ip}$ for any other agent~$i$. In contexts of imperfect
information, agents can only observe the atoms they can act upon. Hence,
an interesting question is whether our reduction of CGS-SPC to CGS-EPC
goes through even when imperfect information is assumed.

\section*{Acknowledgements}

The authors are grateful to the three anonymous reviewers for their
helpful comments.  F.~Belardinelli acknowledges the support of the
French ANR JCJC Project SVeDaS (ANR-16-CE40-0021).

\bibliography{cgs}

\end{document}